\def\year{2021}\relax
\documentclass[letterpaper]{article} 
\usepackage{aaai21}  
\usepackage{times}  
\usepackage{helvet} 
\usepackage{courier}  
\usepackage[hyphens]{url}  
\usepackage{graphicx} 
\usepackage{csquotes}
\usepackage{amssymb}
\usepackage{rotating}
\usepackage{amsmath}
\usepackage{natbib}  
\usepackage{caption} 
\usepackage{array}
\usepackage{setspace} 
\usepackage{makecell}
\usepackage[flushleft]{threeparttable}
\usepackage{booktabs,caption}
\usepackage[toc,page]{appendix}
\usepackage{subcaption}
\usepackage{rotating}
\usepackage{multirow}
\usepackage[T1]{fontenc}

\usepackage[usenames,dvipsnames]{color}
\usepackage[dvipsnames]{xcolor}

\newcommand{\ie}{\textit{i}.\textit{e}., }
\newcommand{\eg}{\textit{e}.\textit{g}. }

\title{Bias or Diversity? Unraveling Fine-Grained Thematic Discrepancy in U.S. News Headlines}

\author {
    Jinsheng Pan\textsuperscript{*}, Weihong Qi\textsuperscript{*}, Zichen Wang, Hanjia Lyu, Jiebo Luo\\
}
\affiliations {
    University of Rochester\\
    \{jpan24,wqi3,zwang189,hlyu5\}@ur.rochester.edu, jluo@cs.rochester.edu\\
     * these authors contributed equally
}

\begin{document}

\maketitle

\begin{abstract}
There is a broad consensus that news media outlets incorporate ideological biases in their news articles. However, prior studies on measuring the discrepancies among media outlets and further dissecting the origins of thematic differences suffer from small sample sizes and limited scope and granularity.  In this study, we use a large dataset of 1.8 million news headlines from major U.S. media outlets spanning from 2014 to 2022 to thoroughly track and dissect the fine-grained thematic discrepancy in U.S. news media. We employ multiple correspondence analysis (MCA) to quantify the fine-grained thematic discrepancy related to four prominent topics - domestic politics, economic issues, social issues, and foreign affairs in order to derive a more holistic analysis. Additionally, we compare the most frequent $n$-grams in media headlines to provide further qualitative insights into our analysis. Our findings indicate that on domestic politics and social issues, the discrepancy can be attributed to a certain degree of media bias. Meanwhile, the discrepancy in reporting foreign affairs is largely attributed to the diversity in individual journalistic styles. Finally, U.S. media outlets show consistency and high similarity in their coverage of economic issues.
\end{abstract}

\section{Introduction}
 News media plays a vital role in influencing public perceptions of domestic politics, economic policies, social issues, and foreign affairs~\citep{soroka2003, LinosKaterina2016, hitt2018, lyu2023computational}. While diverse individual perspectives in news articles promote informed discussions and critical thinking, systematic bias can result in misinformation and heightened polarization of views~\citep{entman2007framing, prior2013media, mourao2019fake,lyu2022understanding}. Understanding the dynamics of fine-grained thematic variations and identifying the underlying causes of such discrepancies provide valuable insights into the media landscape, which is essential for the function of democracy. Fine-grained thematic discrepancy in news media refers to differences in the specific topics selected for the content of news coverage. For instance, when reporting on news related to abortion, a focus on abortion rights and a focus on abortion laws represent two distinct subtopics, even though they are both related to the general theme of abortion. Although existing literature has extensively studied various types of media bias~\citep{d2000media}, media use selectivity~\citep{iyengar2009red, knobloch2020confirmation}, and the consequences of media bias~\citep{jamieson2008echo}, comprehensive research on the landscape of news media topics and whether the thematic differences are attributable to \textbf{media bias or perspective diversity} remains scarce. In this study, we use 1.8 million news headlines from nine U.S. national news media outlets and perform multiple correspondence analysis (MCA) to compute thematic similarity, exploring the American media landscape. We subsequently examine the thematic variations over time and between media outlets to uncover the underlying factors contributing to these differences. 

\textit{Does the thematic discrepancy stem from media bias or diversity in perspectives?} Existing literature presents conflicting arguments and evidence. While a substantial body of research agrees on the existence of ideological biases among the U.S. news media~\citep{sutter2000can,groseclose2005measure, gentzkow2010drives}, \citet{budak2016fair} discover considerable similarities among major outlets, except for political scandals. The conflicting evidence may be due to the different study samples they focus on. For example, the study samples of \citet{d2000media} are news articles about presidential elections, while the study samples of \citet{budak2016fair} are general political articles. Additionally, different methods, such as meta-analysis and machine learning models based on crowd-sourced labels, can yield different results. To provide a holistic understanding of such thematic discrepancy, we extend the scope to \textbf{four prominent topics across multiple national media organizations ranging from 2014 to 2022 at a larger scale}. Three of the topics are domestic politics, social issues, and foreign affairs. \citet{lyu2023computational} highlight the importance of these topics in the assessment of hyperpartisanship across different media. We further include economic issues because this topic also involves different perceptions despite its objectivity~\cite{AngZoe2022}. Instead of full news articles, we concentrate on analyzing the thematic discrepancies in \textbf{news headlines} because they are more accessible and they frequently encapsulate the key opinions or events of the content.  In addition, the headlines achieve an optimal balance between contextual impact and cognitive effort, effectively guiding readers to construct a coherent interpretation of the information presented, as confirmed by~\citet{dor2003newspaper}.

To distinguish the media bias and the perspective diversity, we follow existing literature in defining media bias as 1) selecting and framing particular issues with ideological leaning, 2) distortion of facts, or 3) only reporting negative news about certain parties or ideologies~\cite{d2000media, budak2016fair, gentzkow2010drives}.

\section{Related work}
Despite their role in democratic supervision, news reports may not be free of bias. For instance, \citet{bourgeois2018selection} find selection biases in the context of news coverage. Although the definition of media bias varies, it is widely agreed that \textbf{selecting and framing particular issues with ideological leaning, distortion of facts, and only reporting negative news about certain parties or ideologies are typical types of media bias}~\citep{d2000media, budak2016fair, gentzkow2010drives}. By this definition, political partisan bias, which strategically manipulates headlines, article sizes, and framings to make reports consistent with their ideology is widespread in news media~\cite{groeling1998, Groseclose2005, Groeling2013, Doron20}. However, the thematic discrepancy in news articles does \textbf{not} necessarily attribute to media bias. For example, the different interpretations of the same event, the unique narrative style of individual journalists, and the different individual experiences can all lead to thematic differences in news articles but are not necessarily systematic biases. The aspect of discrepancy is rarely visited by academic scholars. Our study contributes to unraveling the fine-grained thematic variations in U.S. news headlines.

Understanding the thematic discrepancies among media has attracted much attention from the research community. Traditional methods~\citep{GuessAndrew2021, SpindeTimo2021} collect public opinions from different surveys and polls and quantify the media bias into a certain range of values. However, collecting surveys on a large scale is often time-consuming and expensive. Compared to traditional methods, model-based methods are more feasible. Many prior studies~\cite{benamara2007sentiment, Bautin_Vijayarenu_Skiena_2021} have been conducted on measuring media bias from the perspective of sentiment analysis on news headlines. More recent work exploits masked language models to measure semantic discrepancies. For example, \citet{Guo_Ma_Vosoughi_2022} mask the adjacent words of specific bigrams in news article sentences and then use fine-tuned language models to predict the possible words that could fill in the blank. They compare the prediction results to measure the attitudinal difference between media. However, it is noteworthy that pre-trained language models may contain unknown bias from the training corpus~\cite{schramowski2022large}. Our study aims to explore potential thematic discrepancies among media outlets by constructing thematic representations using $n$-grams that are free from pre-training bias.

\section{Material and Method}
In this section, we describe how we collect and preprocess news headlines. We then discuss how we identify the news headlines of the four topics (\ie domestic politics, economic issues, social issues, and foreign affairs). In the end, we detail our approach to analyzing the thematic discrepancy.

\subsection{Data Collection and Preprocessing}
Our study uses the dataset collected by \citet{lyu2023computational}. For the sake of a self-contained paper, we provide a brief overview of the data collection and preprocessing process. To collect data from the news media, they employed two approaches: using the official web API provided by the news media and crawling the web archives and search pages of the news media. They retrieved 1.8 million news headlines from the websites of nine representative media outlets including The New York Times, Bloomberg, CNN, NBC, Wall Street Journal, Christian Science Monitor, The Federalist, Reason, and Washington Times. These media outlets were categorized into three groups: {\tt Left}, {\tt Central}, and {\tt Right} with respect to the political leaning of each media outlet, which is assessed by  \url{allsides.com} and \url{mediabiasfactcheck.com}. More specifically, the {\tt Left} group includes The New York Times, Bloomberg, CNN, and NBC. The {\tt Central} group consists of Wall Street Journal and Christian Science Monitor. The {\tt Right} group contains The Federalist, Reason, and Washington Times. The collected data range from January 2014 to September 2022 covering various topics. They preprocessed the data by performing lemmatization, eliminating stop words, and converting all text to lowercase.

\subsection{Relevant Title Identification}
To identify the news headlines of the four topics, we first search for the most frequent $n$-grams. Following \citet{Guo_Ma_Vosoughi_2022}, we choose to find the most frequent bigrams. By examining each year's data, we have isolated the bigrams that appeared no less than 100 times. In total, we have identified 797 bigrams meeting this criterion. Next, two annotators manually categorize these bigrams into the four relevant topics. Before the annotation, a pilot annotation session where the two annotators read a few sample titles together and discuss the labeling schema is performed. We find that it is easy to label because of the non-ambiguity of the bigrams. For example, (`ukrainian',`refugee') falls under the category of foreign affairs, while (`health', `law') pertains to social issues. Each annotator then labels half of the collected bigrams. Subsequently, we search for titles that contain at least one of these bigrams. Finally, we identify 295,311 news headlines from January 2014 to September 2022 that are related to the four topics. Table~\ref{tab:title_identification} summarizes the number of bigrams and corresponding titles.

\begin{table}[t]
  \begin{center}
    \begin{tabular}{l c c}
    \hline
      \textbf{} & \textbf{\# bigrams} & \textbf{\# news headlines}\\
      \hline
      Foreign affairs & 94 &38,137\\
     
      Domestic politics & 460 & 168,911\\
      
      Economic issue & 116 & 43,576\\
      
      Social issue & 127 & 44,687\\
      \hline
      Total & 797 &295,311 \\\hline
    \end{tabular}
  \end{center}
    \caption{Number of labeled bigrams and collected news headlines for each topic.}
    \label{tab:title_identification}
\end{table}

\subsection{Thematic Discrepancy Analysis}

Although techniques such as text frequent pattern mining~\cite{Han2007FrequentPM} and term-based text clustering~\cite{Aggarwal2012ASO} could be used for text analysis, we find multiple correspondence analysis (MCA)~\cite{hirschfeld_1935} adequate for measuring thematic discrepancy, as demonstrated by \citet{lakhanpal2022sinophobia} who used MCA to investigate textual differences in online hate speech. MCA encodes categorical data and represents them in low-dimensional Euclidean space (\ie 2-D in our study). The thematic discrepancy is calculated as the distance in the low-dimensional space. To perform MCA, we construct a contingency table in which each column represents one of nine media outlets and each row denotes the frequency of occurrence for each $n$-gram in the identified news headlines. To improve robustness, we select the $n$-grams that appear more than 50 times in the tiles of a single media outlet. Bigrams and trigrams are used to construct the contingency table. Unigrams are not included in this study because we observe that in news headlines, most subjects are bigrams or trigrams (\eg names and events). For a more meaningful interpretation, we focus on bigrams and trigrams. Next, we perform singular value decomposition (SVD) to obtain the orthogonal vectors that represent the categorical data. Note that SVD is applied for dimensionality reduction for visualization purpose.

\section{Results}
To investigate the temporal patterns of fine-grained thematic discrepancies in news headlines of different media across various important topics, we employ MCA to analyze the titles of each topic for each year. We visualize the MCA results to reflect the media report discrepancies over time. Each point on the graphs represents a single media outlet, with markers indicating their respective ideological position. To further reveal the dynamics of media report discrepancies, for each topic, we present the top 10 most frequent $n$-grams in 2014, 2018, and 2022 from news headlines of all media outlets, representing the key subjects that these outlets highlight. The three years correspond to the shift in the presidency as well as the pre- and post-COVID eras. We then show the top 10 most frequent $n$-grams for representative outlets from the {\tt Left}, {\tt Central}, and {\tt Right} categories, as well as within and diverging from the majority cluster, to explore the underlying factors contributing to the thematic variation. 

\subsection{Domestic politics}

\begin{figure}[t]
    \centering
    \includegraphics[width=\linewidth]{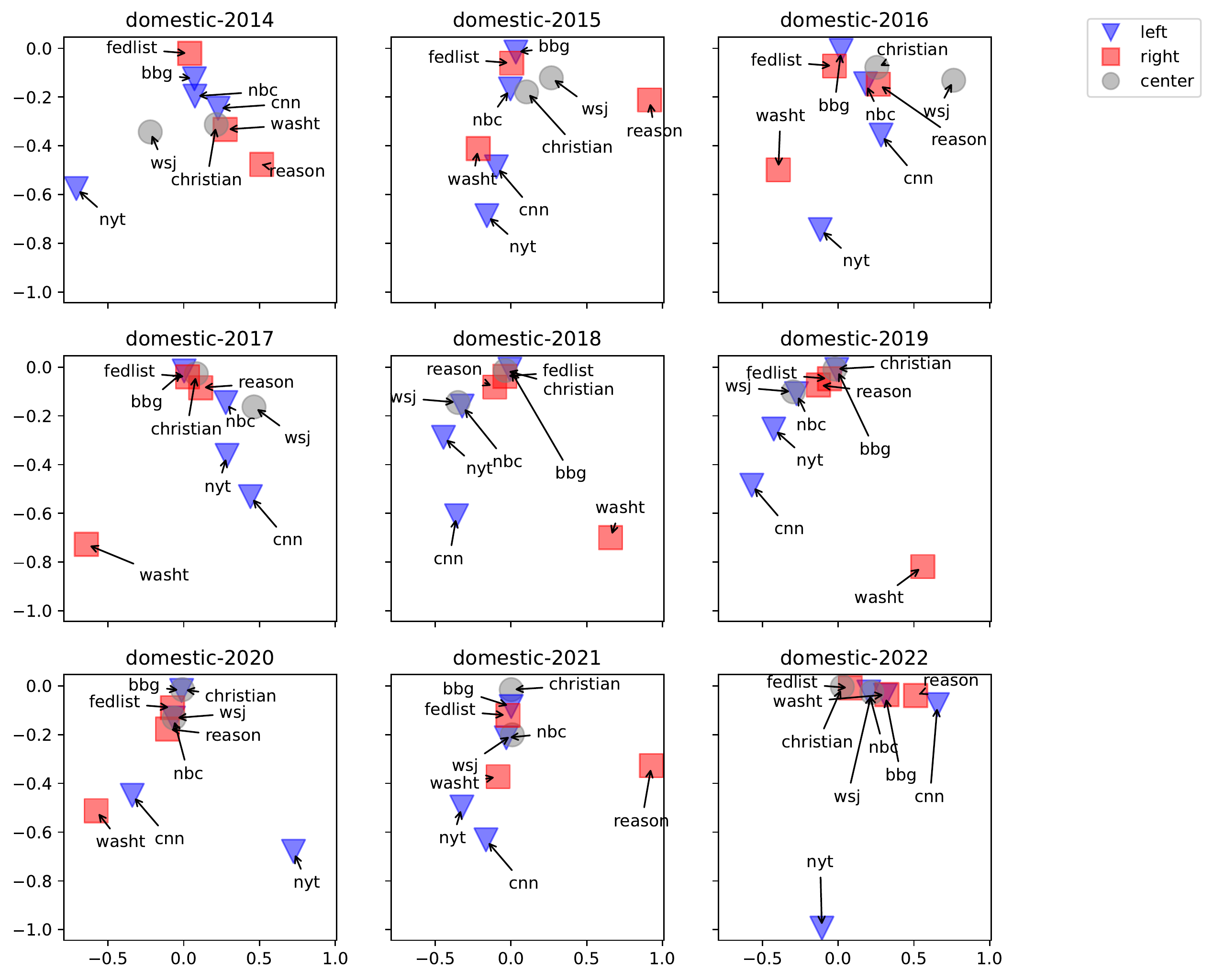}
    \caption{MCA results regarding the coverage of \textit{domestic politics} by the media between 2014 and 2022  (Best viewed by zoom-in on screen).}
    \label{fig:domestic}
\end{figure}

\begin{figure}[t]
    \centering
    \includegraphics[width=\linewidth]{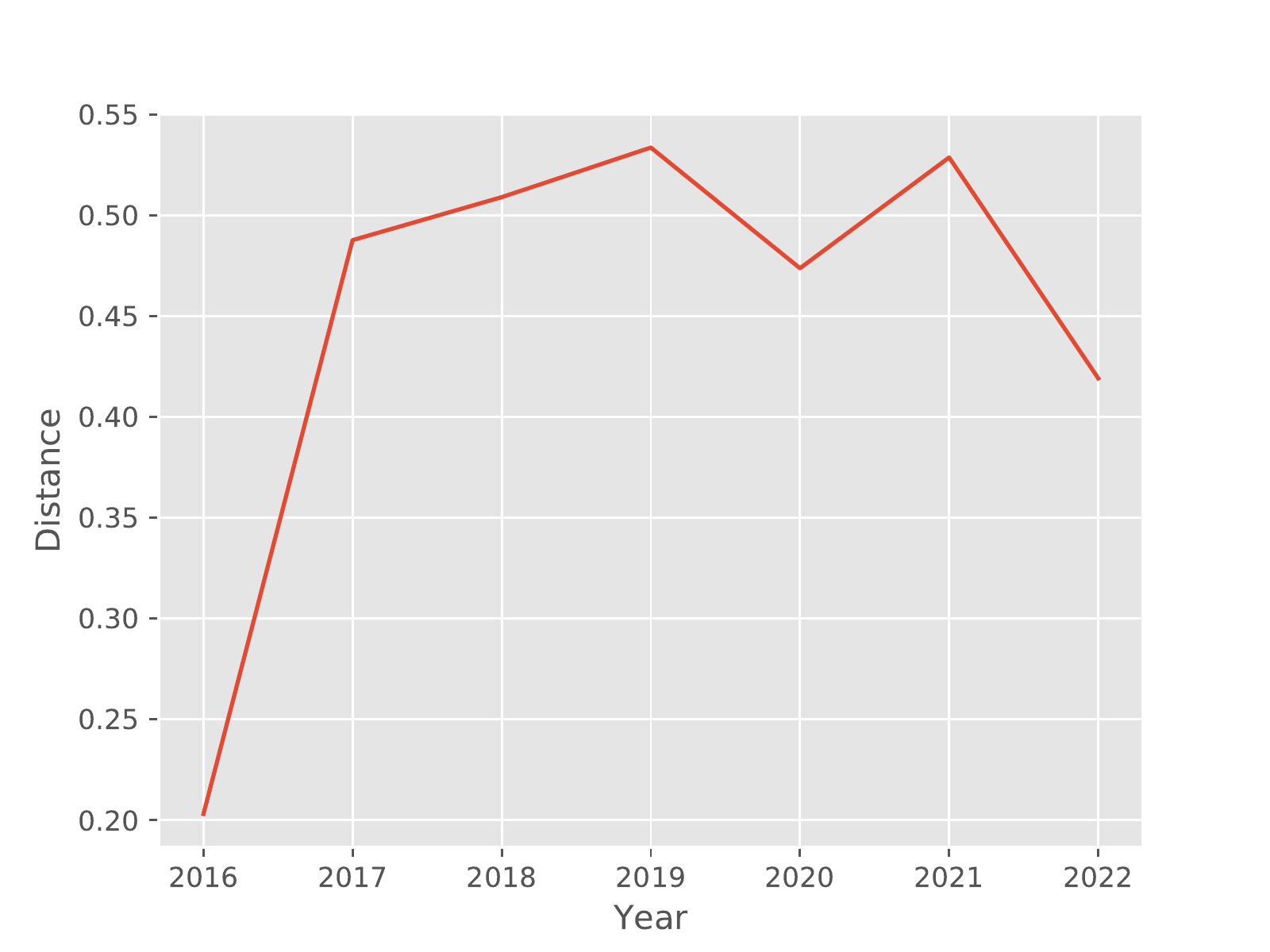}
    \caption{Distance between CNN and the centroid of the major cluster from the MCA results.}
    \label{fig:cnn_vs_major}
\end{figure}

Figure~\ref{fig:domestic} depicts the temporal characteristics of media discrepancies regarding \textit{domestic politics}. Since 2017, we have observed an overall rising level of concentration among the analyzed media outlets. Specifically, these outlets were more sparsely distributed in 2014 but became notably concentrated by 2020, with CNN, Washington Times, and New York Times as outliers. Within the {\tt Left} media, CNN has displayed an increasing discrepancy compared to other outlets since 2015 (Figure~\ref{fig:cnn_vs_major}). Among the {\tt Right} media, Reason exhibits a marked discrepancy during Democratic presidencies, while the discrepancy decreases during the Republican administration. In contrast, The Washington Times displays a contrasting trend, exhibiting greater thematic discrepancies during the Republican presidency and reduced discrepancies during the Democratic administrations. The growing divergence between CNN and other media outlets could be attributed to the 2016 Presidential Election. Previous studies have highlighted the polarizing nature of this election and the controversies surrounding Donald Trump, which have led to subsequent changes in reporting styles across media outlets~\cite{benkler2018network}.

As shown by Table~\ref{tab:domestic_top_10}, political figures (highlighted in \textcolor{Emerald}{green}) consistently attract considerable media attention, occupying two to six positions within the top 10 $n$-grams. In 2018, the $n$-grams representing Trump and his administration were more frequently used. Over time, judicial institutions (highlighted in \textcolor{YellowOrange}{orange}) have continually garnered significant public attention, with the Supreme Court, Attorney General, and Justice Department consistently ranking among the top 10 headline topics. However, civil rights issues can be overshadowed by political upheavals. While free speech and civil rights received substantial attention in 2014, they were later supplanted by politically charged events such as the Capitol Riot and the January $6^{th}$ Committee. \\

Table~\ref{tab:domestic_media} shows the 10 most frequent $n$-grams in news headlines of the Christian Science Monitor, New York Times, and Reason in 2022, representing the {\tt Central}, {\tt Left}, and {\tt Right} media, respectively. Despite their ideological differences, the Christian Science Monitor and Reason cover similar topics, with the New York Times being an outlier in Figure~\ref{fig:domestic}. The New York Times devotes more attention to elections and legislatures (highlighted in \textcolor{YellowOrange}{orange}), but less to political figures (highlighted in \textcolor{Emerald}{green}) compared to the other two outlets, resulting in a greater thematic distance between them. The results suggest that the discrepancy in domestic politics coverage is partly attributable to the choice of varying topics by different media outlets.

\begin{table}[t]
  \begin{center}
  \resizebox{\columnwidth}{!}{
    \begin{tabular}{ccc}
    \hline
      \textbf{2014} & \textbf{2018} & \textbf{2022}\\
      \hline
      white house & \textcolor{Emerald}{donald trump} &\textcolor{YellowOrange}{supreme court} \\
      \textcolor{YellowOrange}{supreme court}& white house & primary election result\\
      \textcolor{Emerald}{bill de blasio} & \textcolor{YellowOrange}{supreme court}  & congressional district\\
      \textcolor{Emerald}{hillary clinton} & \textcolor{Emerald}{trump administration} & \textcolor{Emerald}{biden administration}\\
      \textcolor{Emerald}{rand paul} & 2016 election & january committee\\
      \textcolor{Emerald}{president obama} & \textcolor{Emerald}{president trump} & \textcolor{Emerald}{joe biden}\\
      \textcolor{YellowOrange}{attorney general} & midterm election &\textcolor{YellowOrange}{attorney general}\\
      free speech & \textcolor{YellowOrange}{attorney general} & \textcolor{YellowOrange}{justice department}\\
      civil right  & \textcolor{Emerald}{melania trump} &capitol riot
      \\
      \textcolor{YellowOrange}{justice department}  & \textcolor{Emerald}{hillary clinton} &senate race\\
      \hline
    \end{tabular}}
  \end{center}
    \caption{Top 10 most frequent $n$-grams in 2014, 2018, 2022 regarding \textit{domestic politics}.}
    \label{tab:domestic_top_10}
\end{table}

\begin{table}[t]
  \begin{center}
  \resizebox{\columnwidth}{!}{
    \begin{tabular}{ccc}
    \hline
      \textbf{Christian} & \textbf{NYT} & \textbf{Reason}\\
      \hline
      \textcolor{YellowOrange}{supreme court} & \textcolor{YellowOrange}{election result} &\textcolor{YellowOrange}{supreme court} \\
      \textcolor{Emerald}{sandy hook}& \textcolor{YellowOrange}{primary election} & \textcolor{YellowOrange}{first amendment}\\
      \textcolor{Emerald}{biden sandy}& \textcolor{YellowOrange}{congressional district primary}  &free speech\\
      \textcolor{Emerald}{biden sign} & \textcolor{YellowOrange}{supreme court} & \textcolor{YellowOrange}{court decision}\\
      \textcolor{Emerald}{donald trump} & white house & \textcolor{Emerald}{biden administration}\\
      jan panel & \textcolor{YellowOrange}{governor primary} & \textcolor{Emerald}{kentajin brown jackson}\\
      far right & \textcolor{YellowOrange}{runoff election} & \textcolor{Emerald}{joe biden}\\
      \textcolor{YellowOrange}{overturn roe} & \textcolor{YellowOrange}{attorney general} & \textcolor{YellowOrange}{court reject}\\
      \textcolor{Emerald}{alex jones} & first congressional &capitol riot\\
      right wings  & second congressional &\textcolor{Emerald}{ron desantis}  \\
      \hline
    \end{tabular}
    }
  \end{center}
    \caption{Top 10 most frequent $n$-grams of the Christian Science Monitor, New York Times, and CNN in 2022 regarding \textit{domestic politics}.}
    \label{tab:domestic_media}
\end{table}

\subsection{Economic Issues} 
\begin{figure}[t]
    \centering
    \includegraphics[width=\linewidth]{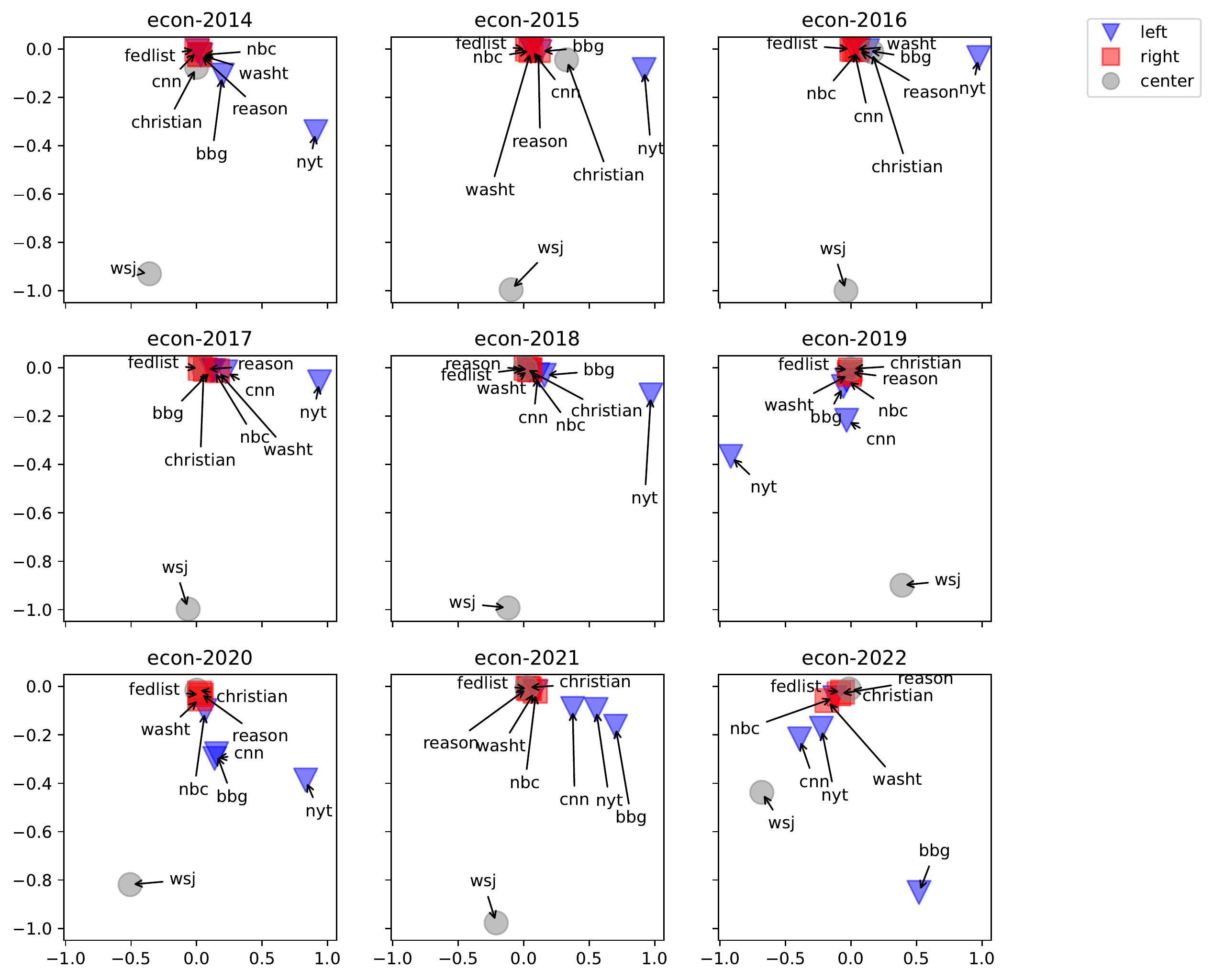}
    \caption{MCA results regarding the coverage of \textit{economic issues} by the media between 2014 and 2022  (Best viewed by zoom-in on screen).}
    \label{fig:economics}
\end{figure}

Figure~\ref{fig:economics} shows the trend of media discrepancies regarding \textit{economic issues} where topic selections within the economic domain exhibits similarities. From 2014 to 2022, the majority of media outlets are situated in the upper left corner of the plot. However, starting in 2019, Bloomberg and CNN have gradually become more distant from the upper left cluster. \\

We have noted minimal changes in the top 10 most frequent $n$-grams in news headlines relating to \textit{economic issues}.\footnote{Details regarding the temporal variation in the most frequent $n$-grams of the economic issues can be found in the Appendix.} We compare the Reason, Wall Street Journal, and CNN, which represent the {\tt Right}, {\tt Central} and {\tt Left} media, respectively. The three outlets exhibit a strong interest in topics including interest rates, the stock market, energy, electronic cars, and essential economic policies. Our findings indicate that the media outlets reveal limited variation in their coverage of economic issues,  and are similar in both topics and perspectives.\\

\begin{table}[t]
  \begin{center}
  \resizebox{\columnwidth}{!}{
    \begin{tabular}{ccc}
    \hline
      \textbf{Reason} & \textbf{WSJ}& \textbf{CNN}\\
      \hline
        student loan forgiveness & stock market &  gas price\\
      gas price& supply chain &  rate hike\\
      biden student debt&  interest rate &  mortgage rate\\
      formula shortage& central bank &  student loan\\
       high gas &  student loan&  interest rate\\
      baby formula shortage&  gas price &  oil price\\
       gas tax&   natural gas & stock market\\
       interest rate &  oil price& natural gas\\
      electric vehicle& real estate &electric car\\
       fossil fuel & electric vehicle & supply chain\\
      \hline
    \end{tabular}
    }
  \end{center}
    \caption{Top 10 most frequent $n$-grams of CNN, Wall Street Journal, and Reason in 2022 regarding \textit{economic issues}. }
    \label{tab:economics_media}
\end{table}

\subsection{Social Issues} 
\begin{figure}[t]
    \centering
    \includegraphics[width=\linewidth]{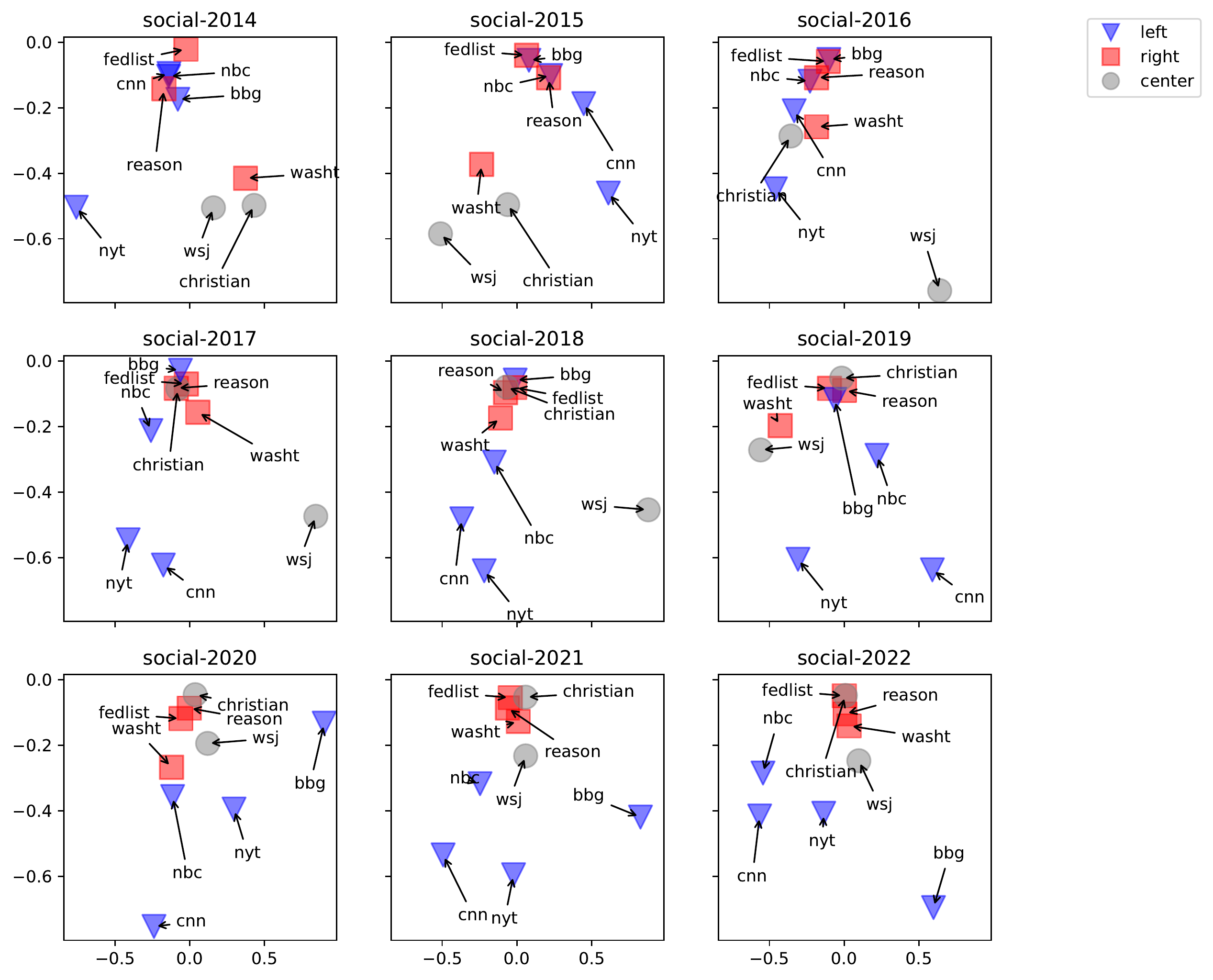}
    \caption{MCA results regarding the coverage of \textit{social issues} by the media between 2014 and 2022  (Best viewed by zoom-in on screen).}
    \label{fig:social}
\end{figure}

Figure~\ref{fig:social} shows the discrepancies among media outlets relating to \textit{social issues}. A similar trend is observed in social issues as in domestic politics, with media outlets becoming more concentrated over time. Some left-wing media outlets, such as CNN, NBC, and Bloomberg, have diverged from the cluster since 2017. Meanwhile, centrist and right-wing media outlets, such as the Christian Science Monitor and Washington Times, have progressively moved closer to the majority. \\

The focus on social issues is likely influenced by ongoing social movements according to the results in Table~\ref{tab:social_top_10}. In 2014, due to the Medicaid expansion~\citep{HimmelsteinGracie2019EotA} and the promotion of same-sex marriage~\citep{nyt-gay}, healthcare and gay marriage (both highlighted in \textcolor{blue}{blue}) obtained significant attention. In 2018, the \#MeToo movement sparked widespread debate and discussion about sexual harassment (highlighted in \textcolor{red}{red}) in the news media~\citep{metoo}. In addition, the overturn of Roe v. Wade in mid-2022~\citep{roe} drew considerable attention to abortion laws and rights (highlighted in \textcolor{Emerald}{green}). \\

Table~\ref{tab:social_media} presents the thematic discrepancies in social issues among the Christian Science Monitor, CNN, and Reason in 2022. We choose the three media outlets to represent the {\tt Central}, {\tt Left}, and {\tt Right} media groups. Notably, the three media outlets cover similar topics including climate change, social media, abortion issues, hate crime, public health, and gun control, but show subtle differences in their approaches and perspectives. For example, while Reason emphasizes ``abortion law'', CNN underscores its ideological position by using ``abortion rights''. Meanwhile, the Christian Science Monitor focuses on ``abortion law'', ``abortion rights'', and ``anti-abortion''. When addressing gun control issues, Reason employs terms like ``mass shooting'', CNN emphasizes ``fatally shoot'', and the Christian Science Monitor uses ``gun violence''. These nuanced differences in the choice of terms reveal the media bias~\cite{d2000media, budak2016fair}.

\begin{table}[t]
  \begin{center}
  \resizebox{\columnwidth}{!}{
    \begin{tabular}{ccc}
    \hline
      \textbf{2014} & \textbf{2018} & \textbf{2022}\\
      \hline
       \textcolor{blue}{health care} & \textcolor{blue}{health care} & climate change\\
      climate change& climate change & social medium\\
      social medium& social medium & mass shooting\\
       \textcolor{blue}{gay marriage}& school shooting & \textcolor{Emerald}{abortion right}\\
       \textcolor{red}{sexual assault}& gun control & hate crime\\
       \textcolor{blue}{health law}& \textcolor{red}{sexual assault} & \textcolor{Emerald}{abortion ban}\\
       \textcolor{blue}{same sex marriage}&  \textcolor{red}{sexual harassment} & \textcolor{blue}{health care}\\
      global warming& \textcolor{red}{sex abuse} &fatally shoot\\
       \textcolor{blue}{health insurance}& \textcolor{red}{sexual misconduct} &school shooting\\
      gun report& gun violence & human right\\
      \hline
    \end{tabular}}
  \end{center}
    \caption{Top 10 most frequent $n$-grams in 2014, 2018, 2022 regarding \textit{social issue}.}
    \label{tab:social_top_10}
\end{table}

\begin{table}[t]
  \begin{center}
    \resizebox{\columnwidth}{!}{
    \begin{tabular}{ccc}
    \hline
      \textbf{Christian} & \textbf{Reason} & \textbf{CNN}\\
      \hline
      climate change & social medium & mass shooting\\
      mass shooting& gun control & social medium\\
      social medium& abortion ban & hate crime\\
      gun violence & public health & climate crisis \\
      abortion law & climate change & fatally shoot \\
      gun control & health care & school shooting \\
      abortion right &  mass shooting & gun violence \\
      hate crime& green energy & uvalde school \\
      green energy &   abortion law & abortion right \\
      \hline
    \end{tabular}}
  \end{center}
    \caption{Top 10 most frequent $n$-grams of the Christian Science Monitor, Reason, and CNN in 2022 regarding \textit{social issues}.}
    \label{tab:social_media}
\end{table}

\subsection{Foreign Affairs}
\begin{figure}[t]
    \centering
    \includegraphics[width=\linewidth]{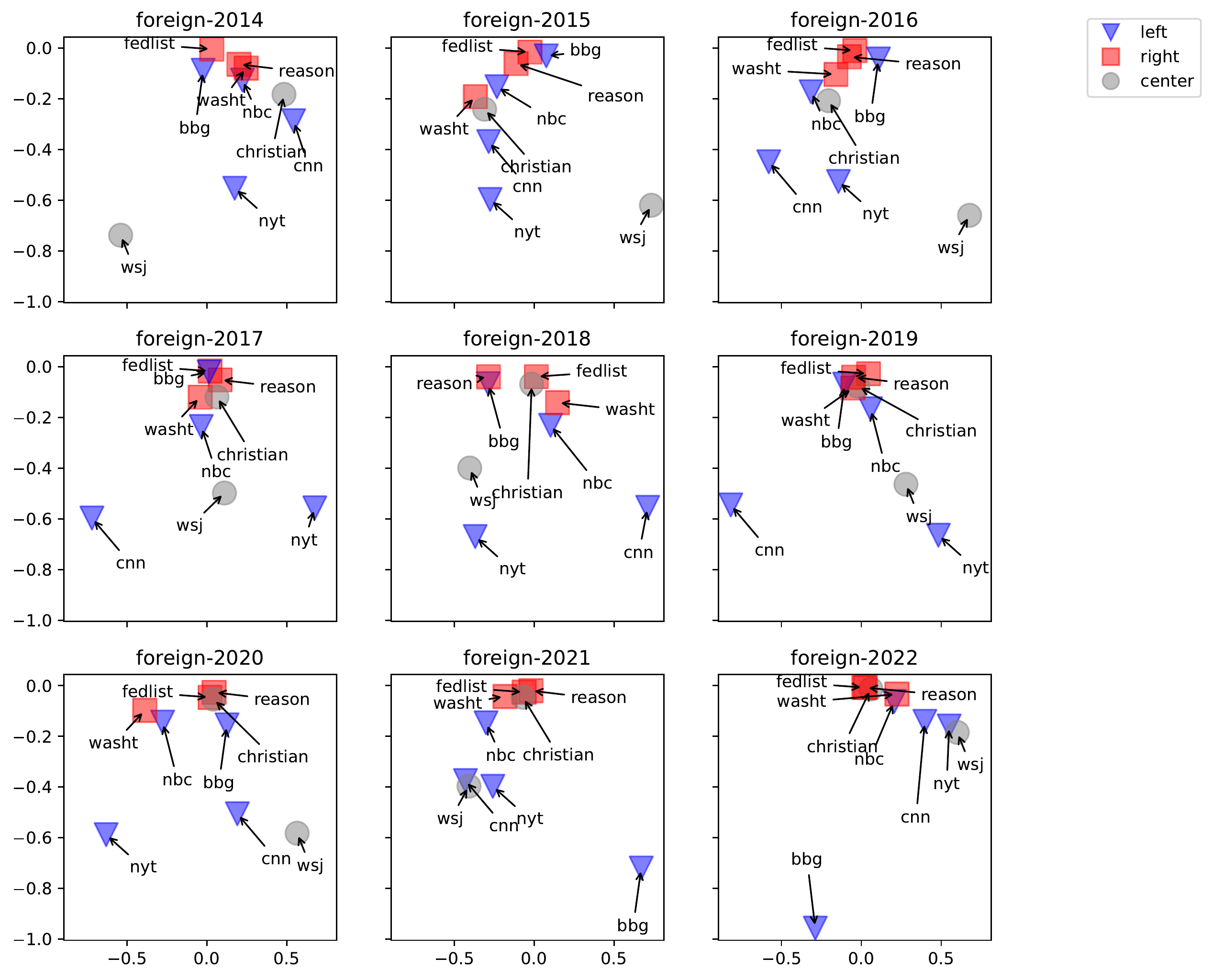}
    \caption{MCA results regarding the coverage of \textit{foreign affairs} by the media between 2014 and 2022  (Best viewed by zoom-in on screen).}
    \label{fig:foreign}
\end{figure}

Figure~\ref{fig:foreign} shows that there has been no significant change in thematic discrepancy relating to \textit{foreign affairs} from 2014 to 2022. The majority of media outlets are clustered in the upper-left corner of the graph, while the New York Times, Wall Street Journal, and CNN emerge as outliers. These three outlets have gradually formed a separate cluster. Additionally, Bloomberg has been distancing itself from the majority since 2020. The emerging outliers may be attributed to the distinct journalistic styles of media outlets. For example, the Wall Street Journal and Bloomberg primarily concentrate on the economic and financial implications of geopolitical tensions, resulting in differing perspectives compared to other media outlets.

The coverage of foreign affairs remains consistent over time, with the exception of war outbreaks.\footnote{Details regarding the temporal variation in the most frequent $n$-grams of the foreign affairs can be found in the Appendix.} Geopolitical tensions have gained the most attention from 2014 to 2022, while the Russia-Ukraine War emerged as a significant topic in 2022. \\

Table~\ref{tab:foreign_media} compares the most frequent $n$-grams of Reason, Wall Street Journal, and Bloomberg in 2022. The three media outlets represent the {\tt Right}, {\tt Central}, and {\tt Left} media. Their coverage of foreign affairs is similar, emphasizing the Russia-Ukraine war, geopolitical tensions, and diplomatic relations with the United Kingdom. Despite the significant discrepancies displayed in Figure~\ref{fig:foreign}, the three media outlets do not exhibit significant differences in word framing within their headlines. This evidence suggests that the discrepancies likely arise from different perspectives and writing styles among journalists, rather than a systematic bias towards a specific ideology.

\begin{table}[t]
  \begin{center}
  \resizebox{\columnwidth}{!}{
    \begin{tabular}{ccc}
    \hline
      \textbf{Reason} & \textbf{WSJ} & \textbf{Bloomberg}\\
      \hline
      ukrainian refugee& ukraine war & hong kong\\
      ukraine war & hong kong & south africa\\
      prime minister & north korea & boris johnson\\
      hong kong & russia ukraine& ukraine war\\
      invasion ukraine & prime minister & south african\\
      russian invasion & boris johnson & saudi arabia\\
      ukraine crisis & russian oil &prime minister\\
      boris johnson & south korea & north korea\\
      middle east  & saudi arabia & russia sanction
      \\
      south africa  & russia sanction & russian oil\\
      \hline
    \end{tabular}}
  \end{center}
    \caption{Top 10 most frequent $n$-grams of Reason, Wall Street Journal, and Bloomberg in 2022 regarding \textit{foreign affairs}.}
    \label{tab:foreign_media}
\end{table}

\begin{figure}[t]
    \centering
    \includegraphics[width=\linewidth]{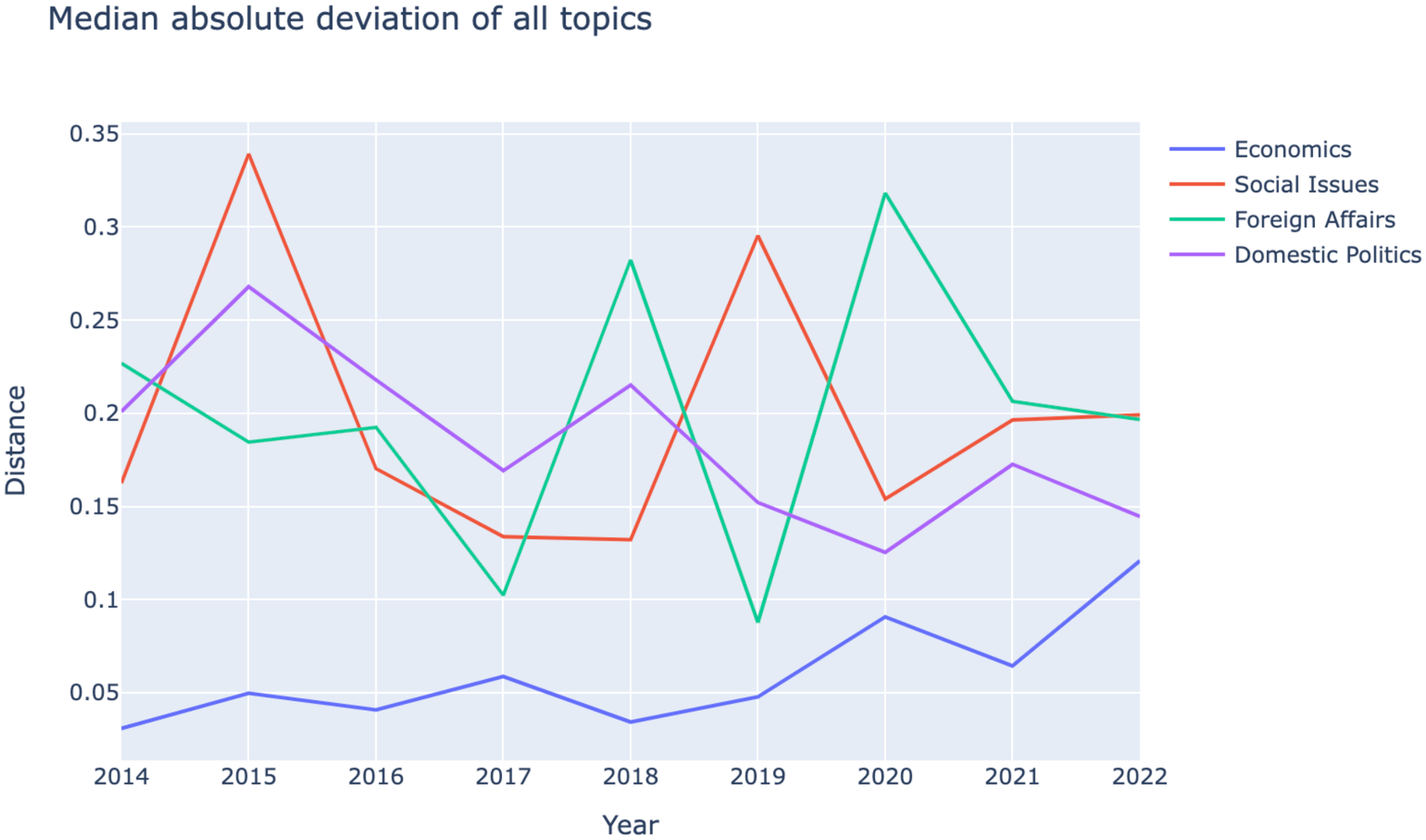}
    \caption{Median absolute deviation of the clusters from the MCA results of \textit{domestic politics}, \textit{economic issues}, \textit{social issues}, and \textit{foreign affairs} between 2014 and 2022.}
    \label{fig:mad}
\end{figure}

\section{Discussions and Conclusions}
News media plays an essential role in the daily lives of the public because it shapes public opinions and attitudes~\cite{brookes2004, LinosKaterina2016, hitt2018}. Different media would have a thematic disagreement on one event due to various reasons, for example, ideology. In this study, we conduct a quantitative and qualitative analysis of the fine-grained thematic discrepancy, focusing on four important topics including domestic politics, economic issues, social issues, and foreign affairs. By performing MCA on the relevant titles of each topic for each year, we discover notable patterns in thematic discrepancies across media in a temporal manner. To quantify the discrepancies, we compute the median absolute deviation of the clusters from the MCA results. The median absolute deviation is used to reduce the sensitivity toward outliers. The results are shown in Figure~\ref{fig:mad}. In regard to \textit{domestic politics}, we find that thematic difference has been diminishing over time. The thematic difference in domestic politics can be mainly attributed to the selection of topics in news coverage.  When it comes to \textit{economic issues}, American news media show little discrepancy both over time and across different outlets. Such consistency of topic preferences also appears in the headlines relating to \textit{foreign affairs}. However, the thematic discrepancy of \textit{foreign affairs} is larger than that of \textit{economic issues}. This pattern indicates that thematic discrepancies may arise from variations in perspectives and writing styles. With respect to \textit{social issues}, the thematic discrepancy is brought about by a mixture of media bias and attitudinal difference. Bias is revealed by different descriptions of the same events and the attitudinal difference is indicated by the topic preference of different media outlets. \\

There are, however, some limitations to our study. First, the media outlets we select are major national news organizations. The gatekeeper and fact-checking mechanisms developed by these outlets help prevent gross misinformation and overt bias, which may result in minimal discrepancies in some issues. Future research could compare national and local media, offering a more comprehensive view of the American news landscape. Second, we only focus on four prominent topics featured in news headlines. To expand upon this, future work could conduct a more fine-grained investigation across a broader range of subjects.

\section{Acknowledgments}
This work was partially supported by the Goergen Institute for Data Science at the University of Rochester.

\bibliography{report} 

\begin{thebibliography}{39}
\providecommand{\natexlab}[1]{#1}
\providecommand{\url}[1]{\texttt{#1}}
\providecommand{\urlprefix}{URL }
\expandafter\ifx\csname urlstyle\endcsname\relax
  \providecommand{\doi}[1]{doi:\discretionary{}{}{}#1}\else
  \providecommand{\doi}{doi:\discretionary{}{}{}\begingroup
  \urlstyle{rm}\Url}\fi

\bibitem[{Aggarwal and Zhai(2012)}]{Aggarwal2012ASO}
Aggarwal, C.~C.; and Zhai, C. 2012.
\newblock A Survey of Text Clustering Algorithms.
\newblock In \emph{Mining Text Data}.

\bibitem[{Ang et~al.(2022)Ang, Reeves, Rogowski, and Vishwanath}]{AngZoe2022}
Ang, Z.; Reeves, A.; Rogowski, J.~C.; and Vishwanath, A. 2022.
\newblock Partisanship, Economic Assessments, and Presidential Accountability.
\newblock \emph{American journal of political science} 66(2): 468--484.
\newblock ISSN 0092-5853.

\bibitem[{Bautin, Vijayarenu, and Skiena(2021)}]{Bautin_Vijayarenu_Skiena_2021}
Bautin, M.; Vijayarenu, L.; and Skiena, S. 2021.
\newblock International Sentiment Analysis for News and Blogs.
\newblock \emph{Proceedings of the International AAAI Conference on Web and
  Social Media} 2(1): 19--26.
\newblock \doi{10.1609/icwsm.v2i1.18606}.
\newblock
  \urlprefix\url{https://ojs.aaai.org/index.php/ICWSM/article/view/18606}.

\bibitem[{Benamara et~al.(2007)Benamara, Cesarano, Picariello, Recupero, and
  Subrahmanian}]{benamara2007sentiment}
Benamara, F.; Cesarano, C.; Picariello, A.; Recupero, D.~R.; and Subrahmanian,
  V.~S. 2007.
\newblock Sentiment analysis: Adjectives and adverbs are better than adjectives
  alone.
\newblock \emph{ICWSM} 7: 203--206.

\bibitem[{Benkler, Faris, and Roberts(2018)}]{benkler2018network}
Benkler, Y.; Faris, R.; and Roberts, H. 2018.
\newblock \emph{Network propaganda: Manipulation, disinformation, and
  radicalization in American politics}.
\newblock Oxford University Press.

\bibitem[{Bourgeois, Rappaz, and Aberer(2018)}]{bourgeois2018selection}
Bourgeois, D.; Rappaz, J.; and Aberer, K. 2018.
\newblock Selection bias in news coverage: learning it, fighting it.
\newblock In \emph{Companion Proceedings of the The Web Conference 2018},
  535--543.

\bibitem[{Brookes, Lewis, and Wahl-Jorgensen(2004)}]{brookes2004}
Brookes, R.; Lewis, J.; and Wahl-Jorgensen, K. 2004.
\newblock The media representation of public opinion: British television news
  coverage of the 2001 general election.
\newblock \emph{Media, Culture \& Society} 26(1): 63--80.

\bibitem[{Budak, Goel, and Rao(2016)}]{budak2016fair}
Budak, C.; Goel, S.; and Rao, J.~M. 2016.
\newblock Fair and balanced? Quantifying media bias through crowdsourced
  content analysis.
\newblock \emph{Public Opinion Quarterly} 80(S1): 250--271.

\bibitem[{D'Alessio and Allen(2000)}]{d2000media}
D'Alessio, D.; and Allen, M. 2000.
\newblock Media bias in presidential elections: A meta-analysis.
\newblock \emph{Journal of communication} 50(4): 133--156.

\bibitem[{Dor(2003)}]{dor2003newspaper}
Dor, D. 2003.
\newblock On newspaper headlines as relevance optimizers.
\newblock \emph{Journal of pragmatics} 35(5): 695--721.

\bibitem[{Entman(2007)}]{entman2007framing}
Entman, R.~M. 2007.
\newblock Framing bias: Media in the distribution of power.
\newblock \emph{Journal of communication} 57(1): 163--173.

\bibitem[{Gentzkow and Shapiro(2010)}]{gentzkow2010drives}
Gentzkow, M.; and Shapiro, J.~M. 2010.
\newblock What drives media slant? Evidence from US daily newspapers.
\newblock \emph{Econometrica} 78(1): 35--71.

\bibitem[{Groeling(2013)}]{Groeling2013}
Groeling, T. 2013.
\newblock Media Bias by the Numbers: Challenges and Opportunities in the
  Empirical Study of Partisan News.
\newblock \emph{Annual Review of Political Science} 16(1): 129--151.
\newblock ISSN 1094-2939.

\bibitem[{Groeling and Kernell(1998)}]{groeling1998}
Groeling, T.; and Kernell, S. 1998.
\newblock Is network news coverage of the president biased?
\newblock \emph{The Journal of Politics} 60(4): 1063--1087.

\bibitem[{Groseclose and Milyo(2005{\natexlab{a}})}]{groseclose2005measure}
Groseclose, T.; and Milyo, J. 2005{\natexlab{a}}.
\newblock A measure of media bias.
\newblock \emph{The quarterly journal of economics} 120(4): 1191--1237.

\bibitem[{Groseclose and Milyo(2005{\natexlab{b}})}]{Groseclose2005}
Groseclose, T.; and Milyo, J. 2005{\natexlab{b}}.
\newblock A Social-science Perspective on Media Bias.
\newblock \emph{Critical review (New York, N.Y.)} 17(3-4): 305--314.
\newblock ISSN 0891-3811.

\bibitem[{Guess et~al.(2021)Guess, Barberá, Munzert, and
  Yang}]{GuessAndrew2021}
Guess, A.; Barberá, P.; Munzert, S.; and Yang, J. 2021.
\newblock The consequences of online partisan media.
\newblock \emph{Proceedings of the National Academy of Sciences} 118:
  e2013464118.
\newblock \doi{10.1073/pnas.2013464118}.

\bibitem[{Guo, Ma, and Vosoughi(2022)}]{Guo_Ma_Vosoughi_2022}
Guo, X.; Ma, W.; and Vosoughi, S. 2022.
\newblock Measuring Media Bias via Masked Language Modeling.
\newblock \emph{Proceedings of the International AAAI Conference on Web and
  Social Media} 16(1): 1404--1408.
\newblock \doi{10.1609/icwsm.v16i1.19396}.
\newblock
  \urlprefix\url{https://ojs.aaai.org/index.php/ICWSM/article/view/19396}.

\bibitem[{Han et~al.(2007)Han, Cheng, Xin, and Yan}]{Han2007FrequentPM}
Han, J.; Cheng, H.; Xin, D.; and Yan, X. 2007.
\newblock Frequent pattern mining: current status and future directions.
\newblock \emph{Data Mining and Knowledge Discovery} 15: 55--86.

\bibitem[{Himmelstein(2019)}]{HimmelsteinGracie2019EotA}
Himmelstein, G. 2019.
\newblock Effect of the Affordable Care Act's Medicaid Expansions on Food
  Security, 2010-2016.
\newblock \emph{American journal of public health (1971)} 109(9): 1243--1248.
\newblock ISSN 0090-0036.

\bibitem[{Hirschfeld(1935)}]{hirschfeld_1935}
Hirschfeld, H.~O. 1935.
\newblock A Connection between Correlation and Contingency.
\newblock \emph{Mathematical Proceedings of the Cambridge Philosophical
  Society} 31(4): 520–524.
\newblock \doi{10.1017/S0305004100013517}.

\bibitem[{Hitt and Searles(2018)}]{hitt2018}
Hitt, M.~P.; and Searles, K. 2018.
\newblock Media coverage and public approval of the US Supreme Court.
\newblock \emph{Political Communication} 35(4): 566--586.

\bibitem[{Iyengar and Hahn(2009)}]{iyengar2009red}
Iyengar, S.; and Hahn, K.~S. 2009.
\newblock Red media, blue media: Evidence of ideological selectivity in media
  use.
\newblock \emph{Journal of communication} 59(1): 19--39.

\bibitem[{Jamieson and Cappella(2008)}]{jamieson2008echo}
Jamieson, K.~H.; and Cappella, J.~N. 2008.
\newblock \emph{Echo chamber: Rush Limbaugh and the conservative media
  establishment}.
\newblock Oxford University Press.

\bibitem[{Knobloch-Westerwick, Mothes, and
  Polavin(2020)}]{knobloch2020confirmation}
Knobloch-Westerwick, S.; Mothes, C.; and Polavin, N. 2020.
\newblock Confirmation bias, ingroup bias, and negativity bias in selective
  exposure to political information.
\newblock \emph{Communication Research} 47(1): 104--124.

\bibitem[{Lakhanpal et~al.(2022)Lakhanpal, Zhang, Li, Lee, Kim, Chae, and
  Kwon}]{lakhanpal2022sinophobia}
Lakhanpal, S.; Zhang, Z.; Li, Q.; Lee, K.; Kim, D.; Chae, H.; and Kwon, H.~K.
  2022.
\newblock Sinophobia, misogyny, facism, and many more: A multi-ethnic approach
  to identifying anti-Asian racism in social media.
\newblock \emph{arXiv preprint arXiv:2210.11640} .

\bibitem[{Linos and Twist(2016)}]{LinosKaterina2016}
Linos, K.; and Twist, K. 2016.
\newblock The Supreme Court, the Media, and Public Opinion: Comparing
  Experimental and Observational Methods.
\newblock \emph{The Journal of legal studies} 45(2): 223--254.
\newblock ISSN 0047-2530.

\bibitem[{Liptak(2014)}]{nyt-gay}
Liptak, A. 2014.
\newblock Justices Reject Call to Halt Gay Marriages in Oregon.
\newblock
  \urlprefix\url{https://www.nytimes.com/2014/06/05/us/politics/supreme-court-rebuffs-call-to-end-same-sex-marriages-in-oregon.html}.

\bibitem[{Liptak(2022)}]{roe}
Liptak, A. 2022.
\newblock In 6-to-3 Ruling, Supreme Court Ends Nearly 50 Years of Abortion
  Rights.
\newblock
  \urlprefix\url{https://www.nytimes.com/2022/06/24/us/roe-wade-overturned-supreme-court.html}.

\bibitem[{Lyu and Luo(2022)}]{lyu2022understanding}
Lyu, H.; and Luo, J. 2022.
\newblock Understanding Political Polarization via Jointly Modeling Users,
  Connections and Multimodal Contents on Heterogeneous Graphs.
\newblock In \emph{Proceedings of the 30th ACM International Conference on
  Multimedia}, 4072--4082.

\bibitem[{Lyu et~al.(2023)Lyu, Pan, Wang, and Luo}]{lyu2023computational}
Lyu, H.; Pan, J.; Wang, Z.; and Luo, J. 2023.
\newblock Computational Assessment of Hyperpartisanship in News Titles.
\newblock \emph{arXiv preprint arXiv:2301.06270} .

\bibitem[{Mour{\~a}o and Robertson(2019)}]{mourao2019fake}
Mour{\~a}o, R.~R.; and Robertson, C.~T. 2019.
\newblock Fake news as discursive integration: An analysis of sites that
  publish false, misleading, hyperpartisan and sensational information.
\newblock \emph{Journalism studies} 20(14): 2077--2095.

\bibitem[{Pomarico(2018)}]{metoo}
Pomarico, N. 2018.
\newblock 11 of the biggest moments of the \#MeToo movement in 2018.
\newblock
  \urlprefix\url{https://www.insider.com/me-too-movement-moments-2018-12}.

\bibitem[{Prior(2013)}]{prior2013media}
Prior, M. 2013.
\newblock Media and political polarization.
\newblock \emph{Annual Review of Political Science} 16: 101--127.

\bibitem[{Schramowski et~al.(2022)Schramowski, Turan, Andersen, Rothkopf, and
  Kersting}]{schramowski2022large}
Schramowski, P.; Turan, C.; Andersen, N.; Rothkopf, C.~A.; and Kersting, K.
  2022.
\newblock Large pre-trained language models contain human-like biases of what
  is right and wrong to do.
\newblock \emph{Nature Machine Intelligence} 4(3): 258--268.

\bibitem[{Shultziner(2020)}]{Doron20}
Shultziner, Doron;~Stukalin, Y. 2020.
\newblock Politicizing What's News: How Partisan Media Bias Occurs in News
  Production.
\newblock \emph{Mass Communication and Society} 24(3): 372--393.
\newblock ISSN 1520-5436.
\newblock \doi{10.1080/15205436.2020.1812083}.
\newblock \urlprefix\url{https://browzine.com/articles/406719000}.

\bibitem[{Soroka(2003)}]{soroka2003}
Soroka, S.~N. 2003.
\newblock Media, public opinion, and foreign policy.
\newblock \emph{Harvard International Journal of Press/Politics} 8(1): 27--48.

\bibitem[{Spinde et~al.(2021)Spinde, Kreuter, Gaissmaier, Hamborg, Gipp, and
  Giese}]{SpindeTimo2021}
Spinde, T.; Kreuter, C.; Gaissmaier, W.; Hamborg, F.; Gipp, B.; and Giese, H.
  2021.
\newblock Do You Think It's Biased? How To Ask For The Perception Of Media Bias
  .

\bibitem[{Sutter(2000)}]{sutter2000can}
Sutter, D. 2000.
\newblock Can the media be so liberal-the economics of media bias.
\newblock \emph{Cato J.} 20: 431.

\end{thebibliography}
\newpage
\appendix
\section{Appendix}
Tables~\ref{tab:economics_top_10} and \ref{tab:foreign_top_10} show the 10 most frequent $n$-grams in 2014, 2018, and 2022 regarding \textit{economic issues} and \textit{foreign affairs}.

\begin{table}[h]
  \begin{center}
  \resizebox{\columnwidth}{!}{
    \begin{tabular}{ccc}
    \hline
      \textbf{2014} & \textbf{2018} & \textbf{2022}\\
      \hline
      small business & central bank & gas price\\
      profit rise& real estate & supply chain\\
      minimum wage& silicon valley & student loan\\
      oil price& stock market & central bank\\
      central bank& oil price& stock market\\
      real estate& government bond & interest rate\\
      natural gas&  interest rate & electric car\\
      silicon valley& u.s government bond &real estate\\
      interest rate& tax cut &russian gas\\
      government bond& supply chain & electric vehicle\\
      \hline
    \end{tabular}}
  \end{center}
    \caption{Top 10 most frequent $n$-grams in 2014, 2018, and 2022 regarding \textit{economic issues}.}
    \label{tab:economics_top_10}
\end{table}

\begin{table}[h]
  \begin{center}
  \resizebox{\columnwidth}{!}{
    \begin{tabular}{ccc}
    \hline
      \textbf{2014} & \textbf{2018} & \textbf{2022}\\
      \hline
      hong kong & north korea & hong kong\\
      north korea& trade war & ukraine war\\
      south korea& saudi arabia & north korea\\
      cease fire& south korea & south africa\\
      foreign policy& kim jong un & boris johnson\\
      prime minister& hong kong & prime minister\\
      ukraine &  north korean &south korea\\
      south africa& prime minister &saudi arabia\\
      south sudan& trump trade&russian oil\\
      middle east& nuclear deal& ukraine invasion\\
      \hline
    \end{tabular}}
  \end{center}
    \caption{Top 10 most frequent $n$-grams in 2014, 2018, and 2022 regarding \textit{foreign affairs}.}
    \label{tab:foreign_top_10}
\end{table}

\end{document}